\documentclass[10pt,twocolumn,letterpaper]{article}
\usepackage{iccv}              
\usepackage{multirow}
\usepackage{ulem} 
\usepackage{amsmath,amssymb,bm}
\definecolor{iccvblue}{rgb}{0.21,0.49,0.74}
\usepackage[pagebackref,breaklinks,colorlinks,allcolors=iccvblue]{hyperref}

\def\paperID{4687} 
\def\confName{ICCV}
\def\confYear{2025}

\newcommand{\model}{VoRA}

\title{Vision as LoRA}

\author{%
  Han Wang$^{1}$ \hspace{1.5cm} 
  Yongjie Ye$^{1}$ \hspace{1.5cm}
  Bingru Li$^{2}$ \hspace{1.5cm}
  Yuxiang Nie$^{1}$ \\ 
  Jinghui Lu$^{1}$ \hspace{1.2cm}
  Jingqun Tang$^{1}$ \hspace{1.2cm}
  Yanjie Wang$^{1}$ \hspace{1.2cm}
  Can Huang$^{1}$ 
  \\
  $^1$ByteDance Inc. \   \ $^2$University of Birmingham
}

\begin{document}
\maketitle
\begin{abstract}
We introduce Vision as LoRA (VoRA), a novel paradigm for transforming an LLM into an MLLM. Unlike prevalent MLLM architectures that rely on external vision modules for vision encoding, VoRA internalizes visual capabilities by integrating vision-specific LoRA layers directly into the LLM. This design allows the added parameters to be seamlessly merged into the LLM during inference, eliminating structural complexity and minimizing computational overhead. Moreover, inheriting the LLM's ability of handling flexible context, \model{} can process inputs at arbitrary resolutions.
\\
To further strengthen VoRA’s visual capabilities, we introduce a block-wise distillation method that transfers visual priors from a pre-trained ViT into the LoRA layers, effectively accelerating training by injecting visual knowledge. Additionally, we apply bi-directional attention masks to better capture the context information of an image. We successfully demonstrate that with additional pre-training data, VoRA can perform comparably with conventional encode-based MLLMs.
All training data, codes, and model weights will be released at \url{https://github.com/Hon-Wong/VoRA}.
\end{abstract}    
\section{Introduction}
\label{sec:intro}

\begin{figure}[ht]
    \centering
    \includegraphics[width=\linewidth]{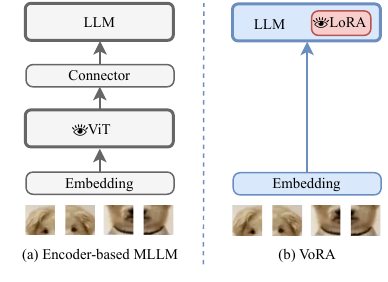}
    \caption{A high-level overview of \model{}. Visual parameters are indicated with an eye icon. Mainstream MLLMs adopt a modular, sequential architecture: raw pixels are first processed by a pre-trained vision encoder to extract high-level visual features, which are then aligned with the LLM through a modality connector for vision-language tasks. In contrast, \model{} consists solely of an LLM and a lightweight embedding layer. The LoRA layers serve as visual parameters that can be integrated into the LLM without incurring additional computational costs or memory burdens. }
    \label{fig:figure1}
\end{figure}

Multimodal Large Language Models (MLLMs) \cite{blip2, llava, llava1_5, flamingo, minigpt} have advanced significantly by integrating pre-trained vision models with Large Language Models (LLMs) \cite{intern1.5, llama, vicuna, gpt3, qwen} through a modular design: visual features extracted by vision encoders to be aligned with LLMs via a connector, as shown in Figure \ref{fig:figure1}(a). While efficient in training, this approach has key limitations derived from the external vision expert models, i.e., extra computational costs and image resolution constraints. For instance, many vision encoders, particularly Vision Transformers (ViTs) \cite{siglip, clip, vit}, adhere to a fixed-resolution training paradigm, limiting flexibility. Additionally, the modular design imposes a sequential workflow: the LLM cannot begin processing until the vision encoder and connector have fully processed the image. To overcome these issues, recent studies \cite{fuyu, eve} have explored unified, encoder-free architectures that process raw pixels directly within a single Transformer (i.e., an LLM), eliminating the need of external vision models. However, such methods face challenges from modality conflicts between vision and language, which would lead to new problems, such as unstable training and catastrophic forgetting issues.

Relevant research \cite{evev2, monointernvl} has made attempts to address modality conflicts through parameter decoupling methods.
For example, Mono-InternVL \cite{monointernvl} introduced a Mixture-of-Experts (MoE) framework \cite{moe}, employing separate expert modules for vision and language processing. Taking a step further, EVEv2 \cite{evev2} decoupled all linear and normalization layers in the LLM. While these approaches helped mitigate modality conflicts, they doubled the LLM’s parameters, complicating the architecture and substantially increasing memory overhead.

To address these challenges, we propose Vision as LoRA (\model{}), a method of transforming LLMs into encoder-free MLLMs by integrating vision understanding abilities through Low-Rank Adaptation (LoRA) \cite{lora}. 
While we acknowledge that decoupling vision and language parameters is critical, we wish to avoid dependency on parameter expansion in inference. To this end, \model{} applies trainable LoRA layers to LLMs, which encode the new modality, i.e., vision, while preserving the language knowledge of the original LLM by freezing its parameters, as shown in Figure \ref{fig:figure1}(b). Unlike previous approaches \cite{evev2, monointernvl} that retain vision-specific parameters during inference, \model{} merges LoRA layers into the LLM  after training, incurring near-zero additional computational cost or memory overhead.

Furthermore, \model{} leverages pre-trained vision models as teacher models to inject visual priors into the LoRA layers. Specifically, we adopt the strategy of block-wise distillation \cite{distillation}: the intermediate visual representations of each LLM block are forced to align with the corresponding block-level features extracted by the teacher model. With such a process, we can greatly accelerate training and reduce the demand for massive data. 

In addition, we replace the LLM's causal attention mask with a bi-directional one for image processing, which better captures contextual relations. Meanwhile, we have also found that, unlike most conventional encoder-based MLLMs \cite{qwen, llava, llava1_5, llavaov, minigpt, minigptv2, cogvlm} which are constrained by fixed-resolution vision encoders, \model{} naturally supports native image resolutions by exploiting the LLM’s inherent ability to process variable-length sequences. 

Our contributions are threefold: 
\begin{itemize}
    \item \textbf{Framework innovation:} \model{} converts LLMs into MLLMs via: (1) vision as LoRA, (2) block-wise distillation, and (3) bi-directional attention for vision. Parameter decoupling between vision and language pathways stabilizes training, while other components accelerate training and reduce data needs. Ablation studies confirm the effectiveness of each element, establishing \model{} as a new paradigm for encoder-free MLLMs.

    \item \textbf{Performance validation:} When trained with a proper scale of additional data, \model{} matches conventional encoder-based MLLMs in terms of performance while reducing computational costs, demonstrating that LLMs can acquire native multimodal capabilities without external vision models. This challenges the widely perceived necessity of encoder-based architectures for multimodal tasks. 

    \item \textbf{Potential extensibility:} Although we narrow down our scope to vision understanding tasks in this paper, the modality-agnostic architecture of \model{} has the potential of generalizing to other modalities (e.g., audio and point clouds) and tasks (e.g., image generation).
\end{itemize}

\section{Related Works}
\label{sec:related}
\subsection{Encoder-based MLLMs}
The dominant architecture of MLLMs has remained largely unchanged since its inception, comprising three components: a ViT \cite{clip, siglip, aimv2}, an LLM \cite{llama, gpt3, qwen2.5, vicuna}, and a connector to bridge modality gaps. Previous research has focused primarily on connector design, ranging from simple MLP layers \cite{llava, llava1_5, minigpt, minigptv2} to hierarchical feature fusion modules \cite{flamingo, llama3.2} or other complex architectures \cite{elysium, dynamicvlm, internvl, cambrian1}. Despite these innovations, fundamental limitations persist due to their reliance on external vision encoders. First, computational and memory overhead escalates dramatically when applying multiple vision encoders \cite{cambrian1} or scaling to larger ones \cite{cogvlm}. Second, fixed-resolution pre-training of ViTs forces MLLMs to employ workarounds like image tiling \cite{llava1_5, llavaov} or restricted square resolutions \cite{cogvlm, qwen}. Recent attempts \cite{pixtral, qwen2vl, aimv2} to train resolution-agnostic ViTs have remained impractical, in that they adopted massive proprietary data and opaque training procedures. These challenges have spurred interest in encoder-free architectures that could bypass ViTs entirely.
\subsection{Encoder-free MLLMs}
The pioneering work, Fuyu \cite{fuyu}, demonstrated the feasibility of training encoder-free models on interleaved image-text data, though at prohibitive computational costs with limited technical transparency. Subsequent approaches, such as EVE \cite{eve}, reduced the vision encoder parameters to a single Transformer block, aligning its output features with a ViT through distillation while updating all LLM parameters to learn about vision during the main training stage. However, these methods still struggle with conflicts between the LLM’s inherent language abilities and the new modality, i.e., vision. These conflicts arise from the coupled language and vision parameters, which exacerbate unstable training and lead to catastrophic forgetting of the original language abilities.

To overcome these problems, Mono-InternVL \cite{monointernvl} and EVEv2 \cite{evev2} proposed parameter decoupling strategies inspired by the MoE method \cite{moe}, duplicating LLM parameters for vision-specific processing while freezing its original weights. Despite successfully addressing forgetting issues and modality conflicts, these methods suffered from substantial memory overhead by doubling model parameters, compromising architectural simplicity. 
Our work addresses this by applying LoRA, which encodes vision while maintaining the language abilities of the LLM, and can be merged into the LLM without causing additional memory overhead.
\section{Vision as LoRA}
\label{sec:methods}
\begin{figure*}
    \centering
    \includegraphics[width=\linewidth]{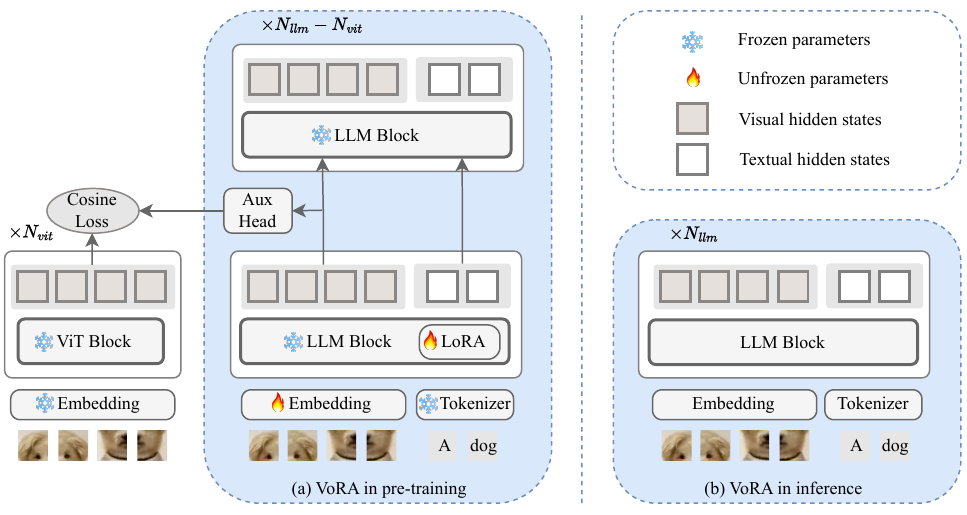}
    \caption{The architecture of \model{}. Figure (a) shows the architecture of \model{} in pre-training: in this stage, \model{} only unfreezes the LoRA layers for vision and the visual embedding layer, i.e., a shallow MLP layer with a positional embedding. Figure (b) shows \model{} in inference: the LoRA layers are merged into the LLM, and thus the only added parameters are a shallow embedding layer (about 6M parameters).}
    \label{fig:architecture}
\end{figure*}
In this section, we introduce three key components of \model{}: vision as LoRA, block-wise distillation, and bi-directional attention masks for vision. 

\subsection{Stabilize training: Vision as LoRA}

As shown in Figure \ref{fig:architecture}(a), we integrate LoRA layers into the LLM to enable vision understanding. During pre-training, images are first converted into vision embeddings using a lightweight embedding layer, i.e., a shallow MLP with positional encodings of about 6M parameters. Let $N_{\text{vit}}$ and $N_{\text{llm}}$ denote the number of blocks in the ViT and the LLM, respectively. We apply LoRA to all linear layers within the first $N_{\text{vit}}$ blocks of the LLM, including query-key-value (QKV) projections and feed-forward network (FFN) layers. Crucially, only the LoRA parameters and the vision embedding layer are updated during training, while the original LLM parameters remain frozen. This design decouples vision and language parameters, stabilizing training compared to full LLM training and avoiding the training collapse observed in prior works \cite{eve}.

Figure \ref{fig:architecture}(b) demonstrates that after pre-training, the LoRA parameters can be seamlessly merged into the base LLM, thereby eliminating additional inference overhead. 

\subsection{Boost training: block-wise distillation}
\label{sec:distillation}

We introduce a block-wise distillation paradigm to align \model{}'s intermediate visual representations with the block-wise features of a pre-trained ViT. This approach transfers visual knowledge from the ViT via knowledge distillation \cite{distillation, eva}, accelerating training while reducing dependence on large-scale vision data. Unlike conventional distillation that updates entire models, we only update the vision-specific LoRA layers within the LLM. Specifically, for each block $i$ in the first $N_{\text{vit}}$ layers of the LLM, we align its hidden states with those of block $i$ in the ViT. 
The training objective combines the following two components.
\\
\textbf{Distillation loss.} For each transformer block $i$ and vision token position $s$, we maximize cosine similarity between projected LLM features and ViT embeddings via:
\begin{equation}
    \mathcal{L}_{\text{distill}}^i = \frac{1}{S} \sum_{s=1}^S \left( 1 - \frac{
        \mathrm{AuxHead}(\bm{h}_{\text{llm}}^{i,s})^\top \bm{h}_{\text{vit}}^{i,s}
    }{
        \|\mathrm{AuxHead}(\bm{h}_{\text{llm}}^{i,s})\|_2 \|\bm{h}_{\text{vit}}^{i,s}\|_2
    } \right),
\end{equation}
where $S$ is the ViT's output sequence length (number of vision embeddings to represent one image), $\bm{h}_{\text{llm}}^{i,s}, \bm{h}_{\text{vit}}^{i,s} \in \mathbb{R}^M$ denote the hidden states for the $s$-th token in block $i$, and $\mathrm{AuxHead}(\cdot)$ is a projection layer (RMSNorm \cite{rmsnorm} + linear layer) adapting LLM features to the ViT's embedding space. The loss is averaged across $N_{\text{vit}}$ blocks:
\begin{equation}
    \mathcal{L}_{\text{distill}} = \frac{1}{N_{\text{vit}}} \sum_{i=1}^{N_{\text{vit}}} \mathcal{L}_{\text{distill}}^i.
\end{equation}
\\
\textbf{Language modeling loss.} For image-caption pairs, we optimize caption generation using cross-entropy, which is consistent with the standard approach used in LLMs:
\begin{equation} 
    \mathcal{L}_{\text{LM}} = -\sum_{t=t_0}^T \log P(w_t | w_{<t}, \bm{x}_{\text{image}}),
\end{equation}
where $T$ is the total sequence length, $\bm{x}_{\text{image}}$ represents vision inputs, and $t_0$ indexes the first caption token.
\\
\textbf{Final objective.} The final loss combines both objectives:
\begin{equation}
    \mathcal{L}_{\text{total}} = \mathcal{L}_{\text{distill}} + \mathcal{L}_{\text{LM}}.
\end{equation}

\subsection{Bi-directional attention masks for vision}

\begin{figure}
    \centering
    \includegraphics[width=\linewidth]{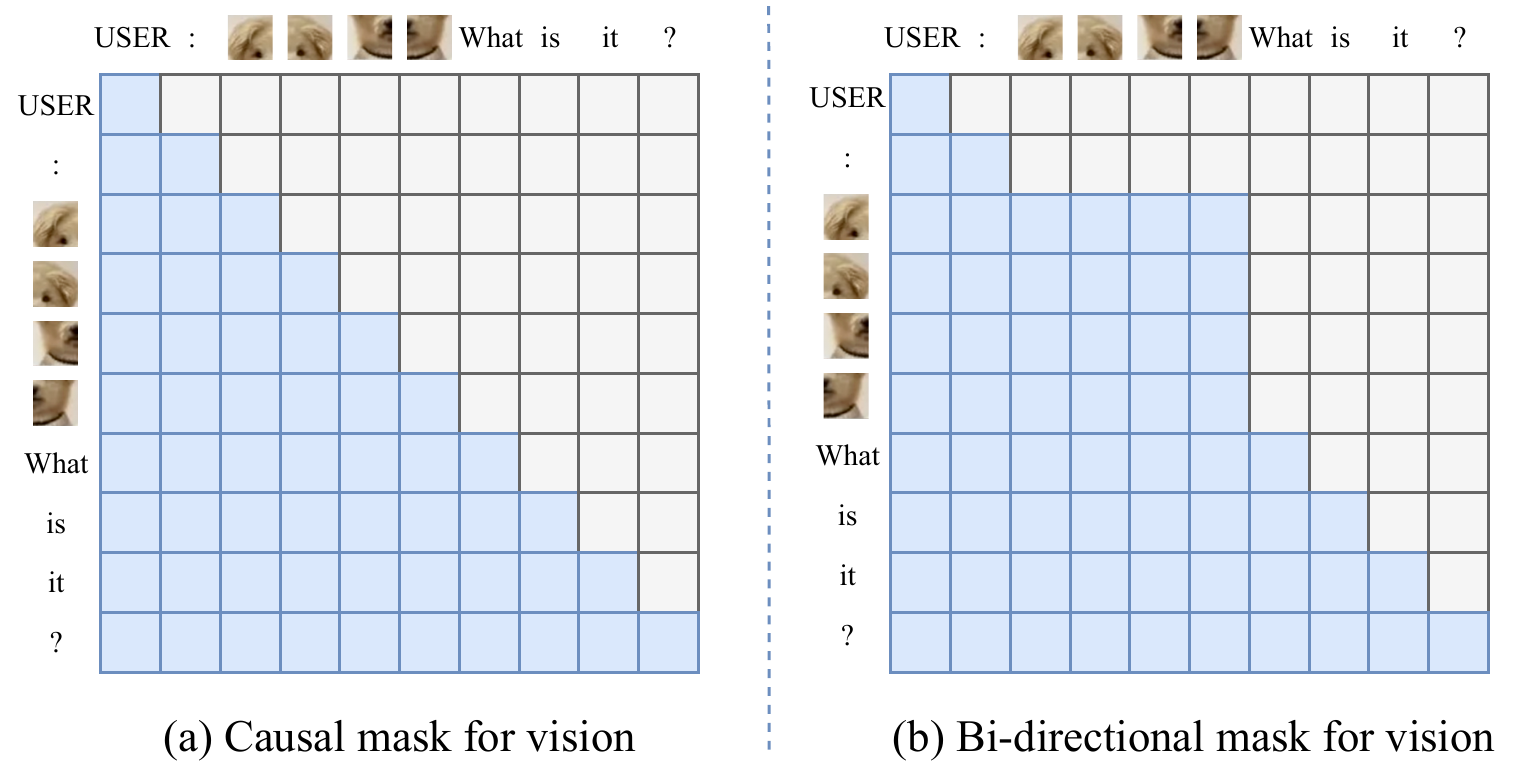}
    \caption{Attention masks for vision: (a) causal attention inherits the autoregressive mask from language modeling, enforcing sequential dependency between image patches; (b) bidirectional attention offers full visibility between all image patches within the same input, enabling global contextual awareness.}
    \label{fig:attention-mask}
\end{figure}

While bi-directional attention masks is common in Transformer architectures in various fields \cite{vit, clip, transfusion}, few studies have explored replacing the causal mask of autoregressive LLMs with a bi-directional mask, especially in the field of MLLMs.

As illustrated in Figure \ref{fig:attention-mask}, we have explored the use of a bi-directional attention mask for vision. Our findings indicate that this attention mask positively impacts the final performance of \model{}, which will be discussed in Section \ref{sec:experiments}. In contrast to prior works \cite{eve, evev2, monointernvl, fuyu}, which have relied on causal masking designed for autoregressive text generation, we demonstrate that adopting bi-directional attention for vision tokens while retaining causal masking for text, not only preserves language capabilities but also enhances visual performance. This aligns with insights from image generation research \cite{transfusion}, highlighting \model{}’s potential as a unified architecture for multimodal generation and understanding tasks.


\section{Data}
\label{sec:data}
\begin{table}[t]
    \centering
    \renewcommand{\arraystretch}{1.2} 
    \setlength{\tabcolsep}{2pt} 
    \small
    \begin{tabular}{l|l|c|c}
        \toprule
         Data Format & Dataset & \# Sample & Total \\
         \midrule
        \multirow{2}{*}{Image Caption} & DataComp29M-recap (ours) & 29M & \multirow{2}{*}{30.4M} \\
         & GLDv2-recap (ours) & 1.4M & \\
        \midrule
        \multirow{10}{*}{Text QA} & Infinity-Instruct-3M \cite{infinityinstruct} & 3.5M & \multirow{10}{*}{6.4M} \\
         & SmolTalk \cite{smoltalk} & 1.0M & \\
         & OpenOrca \cite{OpenOrca} & 994.0K & \\
         & MathInstruct \cite{mathinstruct} & 262.0K & \\
         & OrcaMath \cite{orcamath} & 200.0K & \\
         & MagpiePro (L3 ST) \cite{llavaov} & 150.0K & \\
         & WizardCoder \cite{wizardcoder} & 143.0K & \\
         & OpenCodeInterpreter \cite{opencodeinterpreter} & 66.0K & \\
         & MathQA \cite{mathqa} & 29.8K & \\
         & Dolly \cite{dolly} & 11.0K & \\
        \bottomrule
    \end{tabular}
    \caption{Data used in the pre-training stage of \model{}. We use a mixture of both image and text data to alleviate the forgetting issue in training.}
    \label{tab:data}
\end{table}

\subsection{Data collection and preprocessing}
We claim that the primary focus of this work is not on data engineering or filtration; therefore, we adopt a straightforward data collection and processing strategy. Following previous studies \cite{eve, evev2, monointernvl}, our pre-training framework utilized re-captioned data. Given the limited availability of open-source, large-scale re-captioned datasets, we employed Qwen2-VL-72B \cite{qwen2vl} to generate captions for images sampled from DataComp-1B \cite{datacomp}. From this raw dataset, we selected approximately 29 million images with a longer edge exceeding 448 pixels.

We recognize that this dataset lacks specific world knowledge, particularly regarding landmarks, celebrities, and artworks. To address the deficiency in landmark data, we supplemented our dataset with approximately 1.4 million images from the Google Landmarks Dataset v2 (GLDv2) \cite{googlelandmarksdatasetv2}. For other categories, no suitable million-scale datasets were available. Furthermore, due to potential ethical concerns, we chose not to collect such data. Consequently, we acknowledge that our method may underperform in these domains. However, this limitation can be mitigated in future works by integrating relevant datasets. 

\subsection{Multimodal data mixture}

While \model{} decouples vision and language parameters, we have observed that extended caption-only training slightly degrades the LLM's instruction-following capability. To preserve this ability, we mixed text instruction data into the training data. As shown in Table \ref{tab:data}, our final mixture contained approximately 30M image-caption pairs and 6.4M text instruction samples. The text data were obtained directly from: Infinity-Instruction \cite{infinityinstruct}, SmolTalk \cite{smoltalk}, Cambrian-1 \cite{cambrian1}, and LLaVA-OneVison \cite{llavaov}.
\section{Experiments}
\label{sec:experiments}
\begin{figure}
    \centering
    \includegraphics[width=\linewidth]{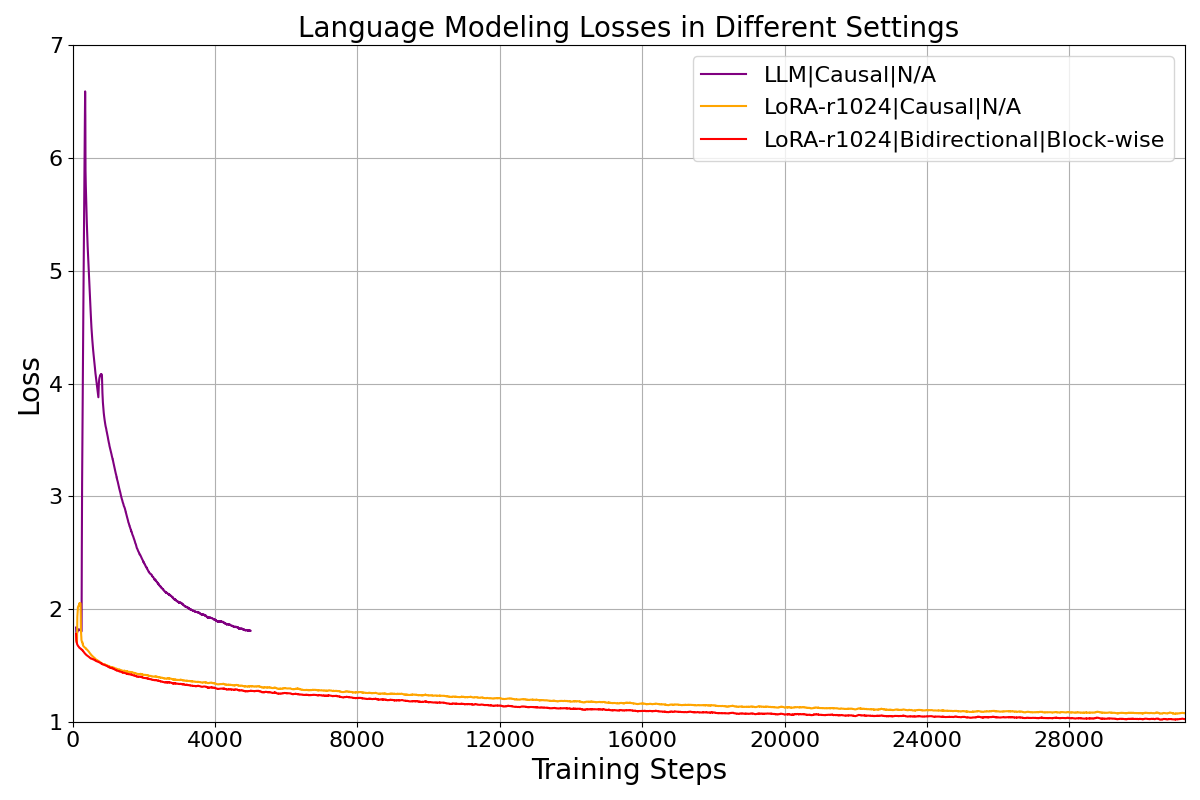}
    \caption{Language modeling losses in different settings. Training the full LLM with a new modality of data can lead to unrecoverable spike in loss curve, i.e., loss collapse.}
    \label{fig:full_llm}
\end{figure}
\subsection{Implementation details}
\textbf{Training setup.}
Unless otherwise specified, we employed AIMv2-Huge-448p \cite{aimv2} as the default vision encoder and Qwen2.5-7B-Instruct \cite{qwen2.5} as the LLM across all experiments. The pre-training learning rate was fixed at 0.0002 (held constant unless explicitly varied), with 100 warm-up steps and a global batch size maintained at 256. All other hyperparameters and optimizer configurations followed the defaults in \cite{llava1_5}. 

For fine-tuning, all LoRA layers were merged into the LLM, while other components (e.g., distillation modules) were eliminated. The full LLM and 6M-parameter visual embedding layer were trainable. For native-resolution variants (\model{}-AnyRes in Table \ref{tab:ablation}), we retained the pre-trained weights of the fixed-resolution version and adopted native-resolution strategy only during fine-tuning.
\\
\textbf{Benchmarks.} As shown in Table \ref{tab:ablation} and Table \ref{tab:mlm_comparison}, we evaluated the model on several benchmarks: VQAv2: VQAv2 \cite{vqav2}; SQA-I: ScienceQA-Image \cite{scienceqa}; TQA: TextVQA \cite{textvqa}; POPE: POPE \cite{pope}; $\mathrm{MMP_p}$: MME Perception \cite{mme}; $\mathrm{MME_c}$: MME Cognition \cite{mme}; MMB: MMBench \cite{mmbench}; SEED-I: SEED-Image \cite{seed}; MMVet: MMVet \cite{mmvet}; AI2D: AI2D \cite{ai2d}; RQA: Realworld-QA \cite{grok1.5v}; MMMU: MMMU \cite{mmmu}.
\subsection{Ablation studies}
\begin{figure*}
    \centering
    \includegraphics[width=\linewidth]{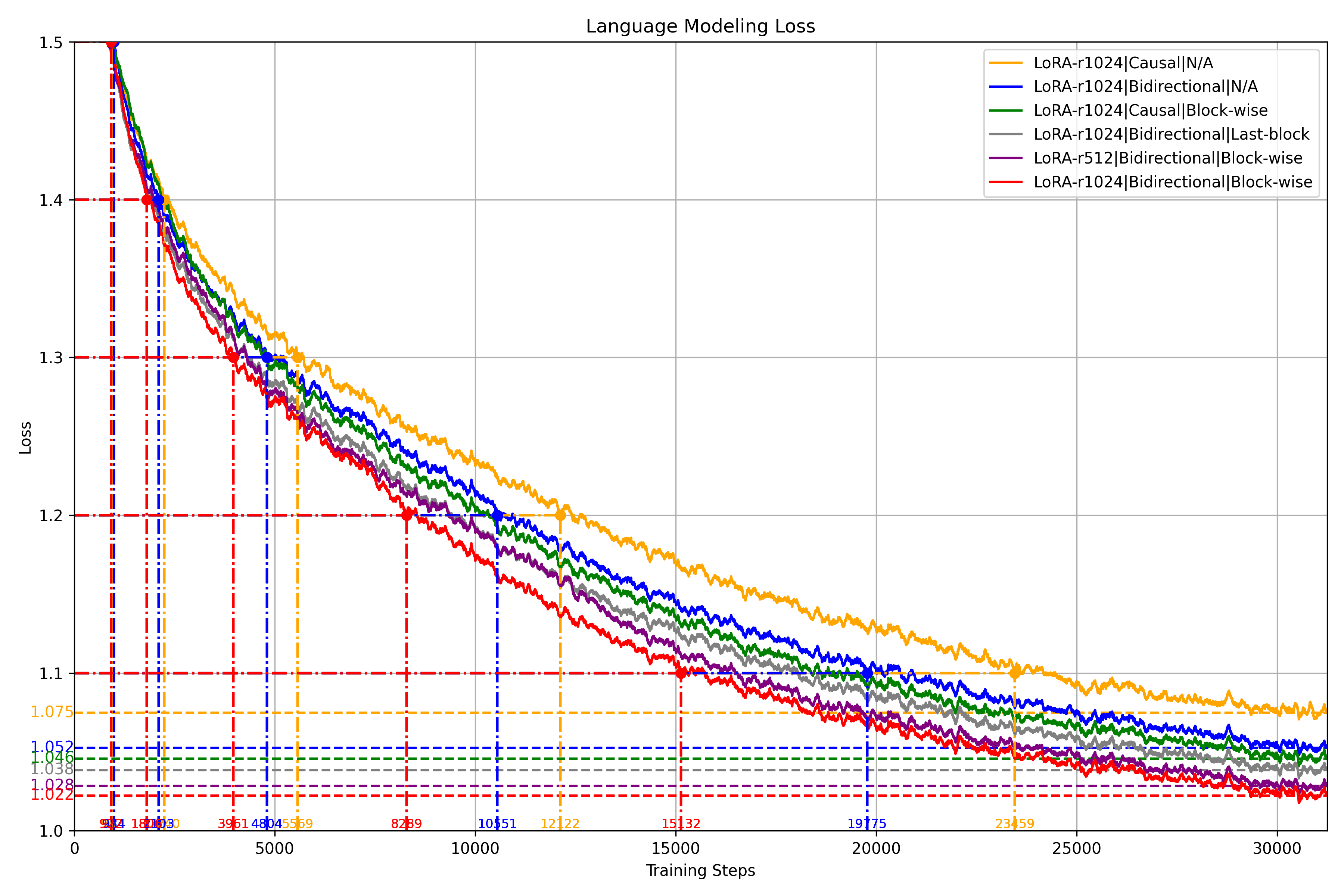}
    \caption{Pre-training loss curves under different configurations. Loss values are smoothed (window=100) for visual clarity. The data sampling order was fixed to ensure fair comparison, as evidenced by the similar trajectories of the loss curves in various settings. LoRA-r1024\textbar Bidirectional\textbar Block-wise refers to the setting: LoRA with rank 1024, bi-directional attention masks for vision, and block-wise distillation. The configuration with the lowest loss was adopted as the default setting in our experiments.}
    \label{fig:ablation_loss}
\end{figure*}
\begin{figure}
    \centering
    \includegraphics[width=\linewidth]{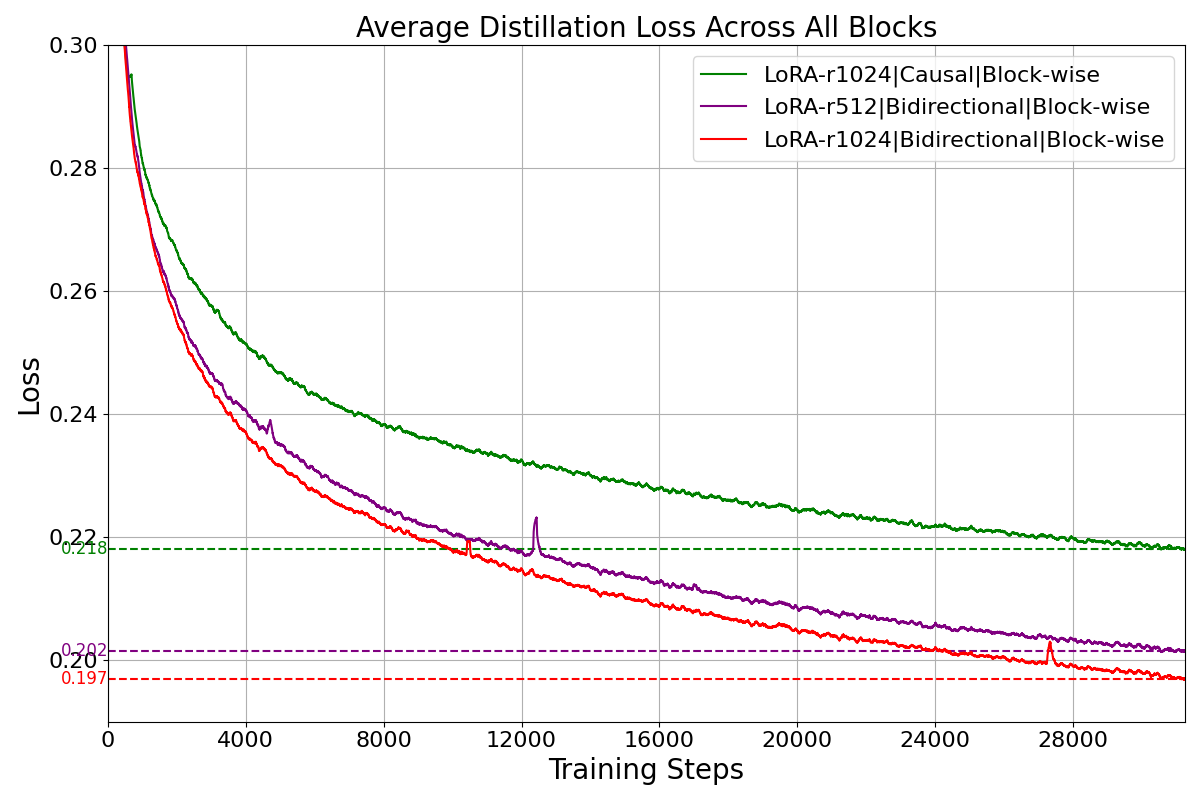}
    \caption{Average distillation loss across all blocks under various settings. Our LoRA-r1024\textbar Bidirectional\textbar Block-wise configuration achieves the lowest average distillation loss across all blocks. This indicates a closer alignment with the ViT’s feature space, confirming that bi-directional attention masks and a larger rank of LoRA layers also enhance visual knowledge transfer.}
    \label{fig:aux_loss}
\end{figure}
\begin{table*}[h]
    \centering
    \renewcommand{\arraystretch}{1.5} 
    \setlength{\tabcolsep}{3.5pt} 
    \small
    \begin{tabular}{l|c|c|cccccccc|c}
    \toprule
    Vision Params & Visual Attention Mask & Distillation type & TQA & POPE & $\mathrm{MME_p}$ & MMB & SEED-I & MMVet & AI2D & RQA & Avg.\\ 
    \midrule
    LoRA-r1024 (2B) & Causal & N.A. & 43.7 & 78.6 & 1137.7 & 47.7 & 57.8 & 20.6 & 49.9 & 49.7 & 50.6 \\
    LoRA-r1024 (2B) & Bidirectional & N.A. & 43.6 & 80.9 & 1132.8 & 49.1 & 58.7 & 17.9 & 47.2 & 51.5 & 50.7 \\
    LoRA-r1024 (2B) & Causal & Block-wise & 45.1 & 82.7 & 1172.9 & 52.9 & 63.7 & 20.1 & 50.9 & 51.2 & 53.2 \\
    LoRA-r1024 (2B) & Bidirectional & Last-block & 44.6 & 82.5 & 1197.5 & 51.8 & 63.8 & 17.9 & 49.9 & 52.8 & 52.9 \\
    LoRA-r512 (1B) & Bidirectional & Block-wise & 47.2 & 83.3 & 1280.5 & 57.6 & 65.3 & 18.5 & 55.9 & 53.1 & 55.6 \\
    LoRA-r1024 (2B) & Bidirectional & Block-wise & 50.1 & 83.8 & 1224.5 & 53.7 & 65.1 & 22.8 & 52.1 & 55.8 & 55.6 \\
    \bottomrule
    \end{tabular}
    \caption{The performance of various settings on standard benchmarks reveals that lower loss during pre-training correlates with better performance. ``LoRA-r1024 (2B)" indicates that the rank for the LoRA layers is set to 1024, with approximately 2 billion parameters unfrozen for training in total.}
    \label{tab:ablation}
\end{table*}
\begin{figure}
    \centering
    \includegraphics[width=\linewidth]{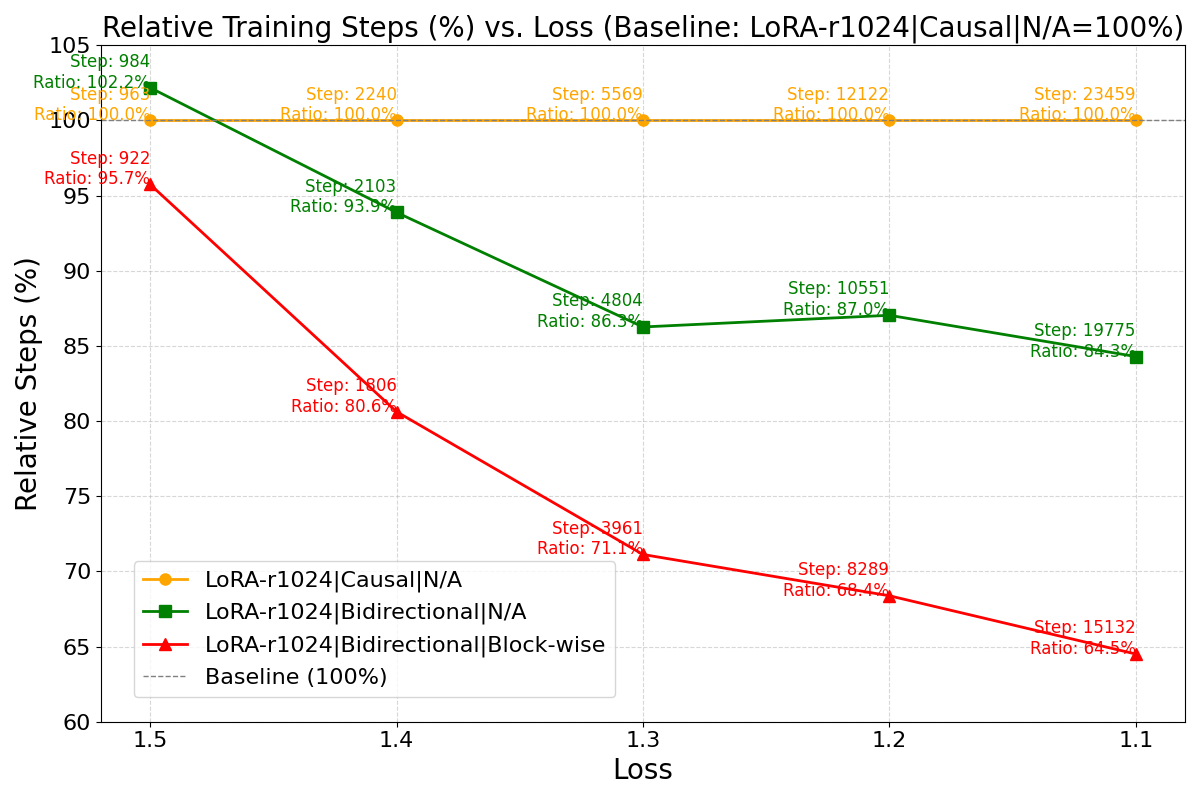}
    \caption{Data efficiency analysis. Our experiments demonstrate that combining bi-directional attention masks for vision tokens with block-wise knowledge distillation significantly improves data efficiency compared to the vanilla LoRA configuration. Furthermore, as the target loss decreases (e.g., from 1.5 to 1.1), the required data proportion relative to the baseline diminishes progressively, indicating higher data efficiency.}
    \label{fig:loss_vs_steps}
\end{figure}
Our ablation studies focused on three key components of \model{}: vision as LoRA, block-wise distillation, and bi-directional attention masks for vision. We employed two primary methods to assess performance in various settings: the pre-training loss on an 8M subset of our DataComp29M-recap dataset, as illustrated in Figure \ref{fig:ablation_loss}, and metrics from eight benchmarks, presented in Table \ref{tab:ablation}. Additionally, we visualized the average distillation loss across all blocks, as shown in Figure \ref{fig:aux_loss}.
\\
\textbf{Ablation on vision as LoRA.} Training the full-parameter LLM proved unstable due to modality conflicts (Figure \ref{fig:full_llm}), consistent with findings in \cite{eve}. While reducing the learning rate to a lower value allowed us to observe one successful training case among several attempts, the loss decreased more slowly than that of LoRA-1024. Therefore, we have excluded it from our primary experiments.

Next, we analyzed different LoRA rank configurations in \model{}. Figure \ref{fig:ablation_loss} shows that a rank of 512 resulted in a slightly higher loss (+0.006) compared to rank 1024. This trend continued in the distillation loss (Figure \ref{fig:aux_loss}), where rank 512 showed a modestly higher average block-wise distillation loss (+0.005) compared to rank 1024. Although both configurations ended up with the same average score of 55.6 (Table \ref{tab:ablation}), the consistent loss advantage suggested that higher ranks might have better optimization potential. Furthermore, we experienced training instability with rank 1536, which prompted us to choose rank 1024 as the default configuration.
\\
\textbf{Ablation on bi-directional attention masks.} 
As demonstrated in Figure~\ref{fig:ablation_loss}, under fixed hyperparameters (e.g., LoRA rank and distillation type), the bi-directional attention mask consistently achieved lower training loss compared to causal masking. This empirical advantage was further supported by the reduced average distillation loss across all Transformer blocks, as depicted in Figure~\ref{fig:aux_loss}. Quantitatively, as evidenced in Table~\ref{tab:ablation}, replacing causal masking with bi-directional masks yielded significant performance improvements. For instance, switching from LoRA-r1024\textbar Causal\textbar Block-wise to LoRA-r1024\textbar Bidirectional\textbar Block-wise led to a 2.4-point average score gain, while replacing LoRA-r1024\textbar Causal\textbar N/A with LoRA-r1024\textbar Bidirectional\textbar N/A yielded a gain of 0.1 points.  
\\
\textbf{Block-wise distillation.} As shown in Figure~\ref{fig:ablation_loss} and Table~\ref{tab:ablation}, applying distillation to the final Transformer block alone significantly improved training efficiency. For example, the transition from the configuration LoRA-r1024\textbar{}Bidirectional\textbar{}N/A to LoRA-r1024\textbar{}Bidirectional\textbar{} Last-block yielded a 2.7-point score gain and a 0.016 reduction in loss. Extending distillation to all blocks via block-wise supervision further enhanced performance: compared with LoRA-r1024\textbar{}Bidirectional\textbar{}Last-block, LoRA-r1024\textbar{}Bidirectional\textbar{}Block-wise produced an additional 2.7-point gain and 0.016 loss reduction. These results indicated that the vanilla distillation method, i.e., last-block distillation, could accelerate training, and block-wise distillation could even strengthen this effect.
\\
\textbf{Data efficiency analysis.} We measured data efficiency by reporting the relative number of training steps required to reach certain loss thresholds, using vanilla LoRA as the baseline. 

As illustrated in Figure~\ref{fig:loss_vs_steps}, the bi-directional attention variant without distillation (LoRA-r1024\textbar{}Bidirectional\textbar{}N/A) required 102.2\% of the baseline training steps to reach Loss=1.5, whereas adding block-wise distillation (LoRA-r1024\textbar{}Bidirectional\textbar{}Block-wise) reduced this to 95.7\%. The efficiency gap became more pronounced at lower loss: at Loss=1.1, the same configurations needed 84.3\% and 64.5\% of the vanilla LoRA baseline steps, respectively. This demonstrated that our optimal configuration achieved equivalent convergence with 35.5\% fewer training steps than vanilla LoRA.

Furthermore, the ratio of data needed by our best configuration relative to vanilla LoRA decreased over time, implying that comparable performance could be achieved with $N\times$ fewer training data.
\subsection{Standard evaluation}
\begin{table*}[h]
    \centering
    \renewcommand{\arraystretch}{1.5} 
    \setlength{\tabcolsep}{0.6pt} 
    \footnotesize 
    \begin{tabular}{llccccccccccccccccc}
        \toprule
        \multirow{2}{*}{Method} & \multirow{2}{*}{LLM} & \multirow{2}{*}{ViT} & \multicolumn{2}{c}{\# Sample} & \multirow{2}{*}{VQAv2} & \multirow{2}{*}{SQA-I} & \multirow{2}{*}{TQA} & \multirow{2}{*}{POPE} & \multirow{2}{*}{$\mathrm{MME_p}$} & \multirow{2}{*}{$\mathrm{MME_c}$} & \multirow{2}{*}{MMB} & \multirow{2}{*}{SEED-I} & \multirow{2}{*}{MMVet} & \multirow{2}{*}{AI2D} & \multirow{2}{*}{RQA} & \multirow{2}{*}{MMMU}\\
        \cmidrule{4-5}
        & & & Pretrain & Finetune & & & & & & & & & & \\
        \midrule
        \textit{Encoder-based}\\
        \midrule
        \rowcolor{gray!30}  
        BLIP2 \cite{blip2} & Vicuna-13B & EVA-1B & 129M & - & 65.0 & 61 & 42.5 & 85.3 & 1293.8 & - & - & 49.7 & 22.4 & - & - & - \\
        \rowcolor{gray!30}  
        InstructBLIP \cite{instructblip} & Vicuna-7B & EVA-1B & 129M & 1.2M & - & 60.5 & 50.1 & - & - & - & 36 & 58.8 & 26.2 & - & - & - \\
        \rowcolor{gray!30}  
        InstructBLIP \cite{instructblip} & Vicuna-13B & EVA-1B & 129M & 1.2M & - & 63.1 & 50.7 & 78.9 & 1212.8 & - & - & - & 25.6 & - & - & - \\
        LLaVA-1.5 \cite{llava1_5} & Vicuna-7B & CLIP-0.3B & 558K & 665K & 78.5 & 66.8 & 58.2 & 85.9 & 1510.7 & 316.1 & 64.3 & 66.1 & 31.1 & 54.8 & 54.8 & 35.3\\
        LLaVA-1.5 \cite{llava1_5} & Qwen2.5-7B & AIMv2-0.6B & 558K & 665K & 82.3 & 77.5 & 59.2 & 85.2 & 1582.3 & 313.0 & 66.3 & 70.6 & 33.7 & 63.7 & 60.0 & 35.3\\
        \midrule
        \textit{Encoder-free}\\
        \midrule
        \model{} & Qwen2.5-7B & \sout{AIMv2-0.6B} & 30M & 665K & 76.0 & 75.9 & 56.3 & 84.5 & 1363.4 & 311.1 & 64.2 & 67.5 & 33.7 & 65.6 & 57.7 & 32.2\\
       \model{}-AnyRes & Qwen2.5-7B & \sout{AIMv2-0.6B} & 30M & 665K & 76.0 & 72.0 & 58.7 & 85.5 & 1336.1 & 319.3 & 61.3 & 68.9 & 33.7 & 61.1 & 60.1 & 32.0\\
       EVE \cite{eve} & Vicuna-7B & \sout{CLIP-0.3B} & 49M(2) & 665K & 75.4 & 63.0 & 51.9 & 83.6 & 1217.3 & 266 & 49.5 & 61.3 & 25.6 & 48.5 & - & - \\
   \rowcolor{gray!30} 
      EVE-HD \cite{eve} & Vicuna-7B & \sout{CLIP-0.3B} & 49M(2) & 1.8M & 74.2 & 64.9 & 56.8 & 85.0 & 1305.7 & 322 & 52.3 & 64.6 & 25.7 & 61.0 & - & - \\
    \rowcolor{gray!30}  
      EVEv2 \cite{evev2} & Qwen2.5-7B & - & 87M(2) & 22.3M(2) & - & 96.2 & 71.1 & 87.6 & - & - & 66.3 & 71.4 & 45.0 & 74.8 & 62.4 & 39.3 \\
  \rowcolor{gray!30}  
      Mono-InternVL \cite{monointernvl} & Intern1.5-2B & - & 1.2B(2) & 150M(2) & - & 93.6 & 72.6 & - & - & - & 65.5 & 67.4 & 40.1 & 68.6 & - & 33.7 \\
      Mono-InternVL \cite{monointernvl} & Intern1.5-2B & - & 922M & 665K & - & 57 & 49 & - & 1100 & - & - & - & - & 42 & - & - \\
    Mono-InternVL \cite{monointernvl} & Intern1.5-2B & - & 1.2B(2) & 665K & - & 58 & 55 & - & 1110 & - & - & - & - & 46 & - & - \\
        \bottomrule
    \end{tabular}
    \caption{Comparison with previous methods on several benchmarks. Since this paper aims to demonstrate that \model{} is a strong base model, we did not scale the fine-tuning data. Therefore, we did not compare with recent state-of-the-art models that often require additional data engineering or involve proprietary datasets; methods that utilize extra fine-tuning data are grayed out. We classified domain-specific VQA data as fine-tuning data rather than pre-training data for EVEv2 and Mono-InternVL, which differs from their original classification in the respective papers. The notation ``49M(2)" indicates that this method employs a two-stage training process using a total of 49M image-text pairs. The strikethrough notation \sout{ViT} means that ViT is excluded during inference.}
    \label{tab:mlm_comparison}
\end{table*}
To ensure a fair comparison between \model{} and existing methods, we deliberately restricted our experimental design. While prior works (e.g., EVE, EVEv2 \cite{evev2}, and Mono-InternVL \cite{monointernvl}) have leveraged massive in-domain datasets (Table~\ref{tab:mlm_comparison}), such approaches complicated direct comparisons due to proprietary training data. Our goal is not to pursue state-of-the-art performance on benchmarks but to validate a novel MLLM architecture. Thus, we limited fine-tuning to the publicly available LLaVA-665K dataset without additional scaling.

To eliminate the potential advantages provided by LLMs and ViTs, we also trained a LLaVA-1.5 model using Qwen-2.5-7B and AIMv2-0.6B. As shown in Table~\ref{tab:mlm_comparison}, prior encoder-free methods often adopted intricate multi-stage pipelines involving module freezing strategies and proprietary datasets (e.g., 100M–1.2B samples). In contrast, our framework employed a streamlined single-stage training process (pre-training followed by fine-tuning), using about 30M image-text pairs. 

As shown in Table~\ref{tab:mlm_comparison}, \model{} achieved performance comparable to both official and reproduced LLaVA-1.5 baselines on most benchmarks when evaluated under strict LLaVA-1.5 protocols \cite{llava1_5}, i.e., identical prompts/generation parameters. However, \model{} underperformed on MME Perception, a gap we attribute to limited world knowledge in our pre-training data. This was further quantified in Table~\ref{tab:world_knowledge}, where \model{} struggled with tasks demanding intensive world-knowledge: 1) inferring movie details from posters, 2) identifying celebrities, 3) recognizing landmarks, and 4) classifying artworks, as these tasks required external domain knowledge absent in our training datasets.

\begin{table}[]
    \centering
    \renewcommand{\arraystretch}{1.5}
    \small
    \setlength{\tabcolsep}{3pt}
    \begin{tabular}{c|cccc|c}
    \toprule
        Method & Posters & Celebrity & Landmark & Artwork & Total\\
        \midrule
        LLaVA-1.5 & 156.1 & 143.5 & 173.5 & 134.0 & 607.1\\
        \midrule
        VoRA & 117.3 & 111.2 & 139.3 & 105.5 & 473.3\\
        VoRA-AnyRes & 110.2 & 104.7 & 138.0 & 110.8 & 463.7\\
    \bottomrule
    \end{tabular}
    \caption{The performance of VoRA in world knowledge tasks. We acknowledge its deficiency, as expected, due to the lack of relevant in-domain data in our pre-training dataset. This is the primary reason for our lower performance on the MME Perception benchmark.}
    \label{tab:world_knowledge}
\end{table}
\section{Limitations}
\label{sec:limitations}

The most significant limitation of \model{} lies in its reliance on additional pre-training data to compensate for the absence of an external vision model, because the LLM has to learn visual feature extraction from scratch.
While we hypothesize that scaling \model{} could surpass encoder-based MLLMs by avoiding information loss in the pre-trained ViT (as theorized in \cite{eve, cambrian1}), we currently lack the empirical evidence to confirm this advantage. Limited training data and computational resources have prevented us from observing a clear performance crossover point. We leave this promising hypothesis for future exploration.

A second limitation is \model{}'s lack of vision token compression. Unlike conventional encoder-based MLLMs that reduce tokens in the connector, we intentionally preserve the original LLaVA-1.5 configurations for fair comparison. This defect could be mitigated in practice through larger patch sizes \cite{eve, evev2, monointernvl} or token pruning/merging techniques. Finally, although \model{} underperformed on world knowledge tasks, this reflects data limitations rather than architectural constraints, and could be resolved through data engineering.

\section{Conclusion and Future Directions}
\label{sec:conclusion}

\model{} establishes a new paradigm for converting LLMs into MLLMs through three components: (1) vision as LoRA, (2) Block-wise distillation, and (3) bi-directional attention masks for vision. 
By integrating vision capabilities directly into the LLM via mergeable LoRA layers for visual encoding, \model{} eliminates the need for a separate vision model. This unified approach reduces memory overhead, lowers computational costs, and leverages the LLM’s inherent flexibility in context length to process native-resolution images with minimal adaptation. This design bypasses the problems brought by using ViT as an external vision model while still decoupling the vision and language parameters to ensure stable training.

The architecture of \model{} is inherently extensible beyond vision-language tasks. By replacing the vision expert model with pre-trained modality-specific experts (e.g., for audio, point clouds, or biomedical signals) and applying new LoRA layers, one can create efficient audio-language, 3D-language, or omni-modal variants without increasing computational costs. 
We envision \model{} as a step toward more unified multimodal intelligence, where a single architecture seamlessly processes diverse modalities and relevant tasks while maintaining inference efficiency.



{
    \small
    \bibliographystyle{ieeenat_fullname}
    \bibliography{main}
}

\end{document}